\documentclass[letterpaper, 10 pt, conference]{ieeeconf}  
\usepackage{graphicx}
\usepackage{multirow}
\usepackage{amssymb}

\usepackage{mathptmx} 
\usepackage{times} 
\usepackage{amsmath} 
\usepackage{bbding}
\usepackage{xcolor}
\usepackage{array}
\newcolumntype{C}[1]{>{\centering\arraybackslash}p{#1}}
\usepackage{makecell}
\usepackage{parskip}
\usepackage{parskip}
\usepackage{url}
\usepackage{multirow}
\usepackage{bm}

\IEEEoverridecommandlockouts                              

\title{\LARGE \bf
Kaninfradet3D:A Road-side Camera-LiDAR Fusion 3D Perception Model based on Nonlinear Feature Extraction and Intrinsic Correlation
}

\author{Pei Liu, Nanfang Zheng, Yiqun Li, Junlan Chen, Ziyuan Pu, ~\IEEEmembership{Member,~IEEE}
\thanks{Manuscript received xxxx, xxxx; revised xxxx, xxxx (\emph {Corresponding thor: Ziyuan Pu})}
\thanks{Pei Liu is with Intelligent Transportation Thrust, Systems Hub, The Hong Kong University of Science and Technology (Guangzhou), No.1 Du Xue Rd, Nansha District, Guangzhou, 511458, China
        {\tt\small pliu061@connect.hkust-gz.edu.cn}}%
\thanks{Nanfang Zheng, Yiqun Li, Junlan Chen and Ziyuan Pu are with School of Transportation, Southeast University, Nanjing, 211102, China
        {\tt\small 220243496@seu.edu.cn; 230248512@seu.edu.cn; junlan.chen@monash.edu; ziyuanpu@seu.edu.cn}}%
}

\everymath{\displaystyle}
\begin{document}

\maketitle
\thispagestyle{empty}
\pagestyle{empty}

\begin{abstract}

With the development of AI-assisted driving, numerous methods have emerged for ego-vehicle 3D perception tasks, but there has been limited research on roadside perception. With its ability to provide a global view and a broader sensing range, the roadside perspective is worth developing. LiDAR provides precise three-dimensional spatial information, while cameras offer semantic information. These two modalities are complementary in 3D detection. However, adding camera data does not increase accuracy in some studies since the information extraction and fusion procedure is not sufficiently reliable. Recently, Kolmogorov-Arnold Networks (KANs) have been proposed as replacements for MLPs, which are better suited for high-dimensional, complex data. Both the camera and the LiDAR provide high-dimensional information, and employing KANs should enhance the extraction of valuable features to produce better fusion outcomes. This paper proposes Kaninfradet3D, which optimizes the feature extraction and fusion modules. To extract features from complex high-dimensional data, the model's encoder and fuser modules were improved using KAN Layers. Cross-attention was applied to enhance feature fusion, and visual comparisons verified that camera features were more evenly integrated. This addressed the issue of camera features being abnormally concentrated, negatively impacting fusion. Compared to the benchmark, our approach shows improvements of +9.87 mAP and +10.64 mAP in the two viewpoints of the TUMTraf Intersection Dataset and an improvement of +1.40 mAP in the roadside end of the TUMTraf V2X Cooperative Perception Dataset. The results indicate that Kaninfradet3D can effectively fuse features, demonstrating the potential of applying KANs in roadside perception tasks.

\end{abstract}

\section{Introduction}

Robust and effective road environment perception is the premise of realizing an automatic driving system. Compared to ego-vehicle perception, roadside perception offers several advantages: an overhead global perspective can avoid target occlusion issues, reduce blind spots in perception, and provide longer-range target detection. Additionally, ego-vehicle perception depends on advanced sensor equipment and computational power on the vehicle, increasing the cost of autonomous vehicles. Roadside perception relies on public infrastructure's sensing equipment and computing power, which is conducive to the future popularization of autonomous driving.

3D road environment perception can rely on various sensors, with cameras and LiDAR being the most fundamental modalities. Camera sensors can provide image data and dense semantic information, while LiDAR can give accurate and dense three-dimensional spatial point cloud positions. Both modalities contain a wealth of information. Currently, many studies focus on camera-LiDAR fusion. Conversely, LiDAR is more effective at figuring out an object's 3D spatial position, while visual data is superior at classifying items. However, not all information provided by LiDAR and cameras is beneficial for training. For example, Zimmer et al. found that using LiDAR data with Coopdet3D and InfraDet3D on the TUMTraf Intersection Dataset resulted in higher mAP compared to the combination of camera and LiDAR \cite{zimmer2024tumtraf}. Besides the final object detection metrics, current research lacks methods to determine whether extracted features are beneficial or reasonable. This study uses a cross-attention block to capture the interdependencies between the two modalities and visualizes the features to demonstrate the useful information extracted from both modalities.

In 3D object detection, transformer modules and convolutional neural networks (CNNs)-based modules are commonly used to extract features from both branches. These modules are limited to linear modeling, which presents challenges in handling complex information. Additionally, as black-box models, they lack interpretability. Recently, Kolmogorov-Arnold Networks (KAN) have provided modules based on combinations of nonlinear functions. These can replace multiple continuous linear layers in the model, resulting in stronger fitting capabilities and improved interpretability. Thus, this study proposes Kaninfradet3D, a roadside radar-vision fusion 3D perception model designed to extract nonlinear features better and enhance fusion performance. In this model, KANs-enhanced encoders are utilized to extract nonlinear features, and the intrinsic correlation between the two modalities is also calculated to improve the fusion process. Our main contributions are:

\begin{itemize}
    \item We proposed improved LiDAR and visual feature encoders based on KANs, addressing the inadequate feature extraction of complex data. Models using these encoders achieved higher accuracy.
    \item We developed the Camera-LiDAR CrossAttn module to address the issue of reduced accuracy after integrating camera features. This module uses attention weights to focus on valuable information. Visual analysis showed that features fused using this module are more meaningful than direct convolution-based fusion.
    \item Our proposed Kaninfradet3D model surpasses the benchmark in detection accuracy, offering a reference for integrating KANs in 3D detection tasks.
\end{itemize}

\section{Related Work}

\subsection{Roadside perception}

At present, the related research of 3D perception mainly focuses on the perspective of the vehicle, and relatively few rely on roadside infrastructures. This is mainly due to the lack of supporting roadside datasets. Ye et al. released Rope3D \cite{ye2022rope3d}, a large-scale roadside dataset containing 3D objects on images. Subsequently, Yu et al. released the large-scale multimodal dataset DAIR-V2X \cite{yu2022dair}, which includes data from both vehicle and roadside perspectives, with DAIR-V2X-I focusing on the roadside perspective and containing LiDAR and image data. Zimmer et al. released the TUMTraf V2X Cooperative Perception Dataset, which also includes both vehicle and roadside data, providing multiple viewpoints of an intersection. These datasets have advanced the development of roadside perception.

BEVHeight \cite{yang2023bevheight} and BEVHeight++ \cite{yang2023bevheight++} are methods that rely solely on vision. They regress the height of vehicles to the ground instead of predicting the pixel-wise depth. Wang et al. proposed the pillar attention fusion network \cite{wang2024pafnet}, which uses LiDAR data from both vehicle and roadside perspectives for 3D detection. However, research on LiDAR and vision fusion is still relatively scarce. Zimmer et al. released InfraDet3D \cite{zimmer2023infradet3d}, which uses early fusion for two LiDARs and late fusion to combine monocular camera information. They then proposed Coopdet3D based on BEVFusion \cite{liu2023bevfusion} and PillarGrid \cite{bai2022pillargrid}, using deep fusion to include roadside camera-LiDAR fusion, vehicle camera-LiDAR fusion, and vehicle- infrastructure fusion in three pipelines. However, the mAP metric for InfraDet3D and Coopdet3D using only LiDAR on TUMTraf Intersection test set \cite{zimmer2023tumtraf} is higher than that for the combination of camera and LiDAR, indicating that more information did not necessarily improve detection performance.

\subsection{Kolmogorov–Arnold Networks (KANs)}

Inspired by the Kolmogorov-Arnold representation theorem, researchers proposed Kolmogorov-Arnold Networks (KANs) \cite{liu2024kan} as promising alternatives for Multi-Layer Perceptrons (MLPs). KANs employ learnable activation functions to eliminate the combination of fixed activation functions and linear functions in MLPs, enabling them to effectively learn high-dimensional complex data. Furthermore, it offers superior interpretability and explainability compared to MLPs. In place of conventional convolutional neural networks (CNNs), Bodner et al. \cite{bodner2024convolutional} presented convolutional Kolmogorov-Arnold networks(Convolutional KANs). In \cite{drokin2024kolmogorov}, Drokin introduced KAN convolutional versions of focal modulation layers and self-attention, as well as a parameter-efficient finetuning procedure for Convolutional KANs. Subsequently, Azam et al.\cite{azam2024suitability} further analyzed the performance and efficiency of constructing traditional building blocks, such as convolutional layers and linear layers, using the KAN concept, highlighting its advantages on visual datasets. However, current research remains limited to image data processing and lacks practical application attempts. This paper explores the application of KANs in the engineering field and attempts to apply KANs to LiDAR data and 3D object detection tasks.

\subsection{Multi-sensor Fusion}

Multi-sensor fusion enables the complementary functionality of data from different modalities, leading to improved performance. The fusion of LiDAR and camera data is one of the most common combinations. Based on the stage of fusion, multi-sensor fusion can be classified into early fusion, deep fusion, and late fusion. Early fusion \cite{gao2018object} refers to directly combining the data before processing, such as aligning and projecting point cloud data to minimize data loss. For example, \cite{8489732} employs LiDAR projections to generate pseudo-images, which are then concatenated with camera images for 2D perception.
Deep fusion, also known as feature-level fusion, involves extracting features from the LiDAR branch, such as voxelizing, before further combining them. MVXNet\cite{sindagi2019mvx}, for instance, projects voxelized features into image space to generate regions of interest (ROIs) and encodes features within these ROIs. \cite{li2022deepfusion} and \cite{Bai_2022_CVPR} incorporate multi-head attention mechanisms to achieve adaptive fusion.
Late fusion \cite{zimmer2023infradet3d} involves merging the proposals from multiple modalities. Its advantage lies in the flexibility to optimize the model for each branch independently.
However, these fusion methods are primarily developed for the ego-vehicle perspective,  and there is still much to explore in developing models that perform fusion from the roadside perspective.

\section{Methodology}
\subsection{KAN-Improved Feature Extraction Module}

\begin{figure*}
    \centering
    \includegraphics[width=0.9\linewidth]{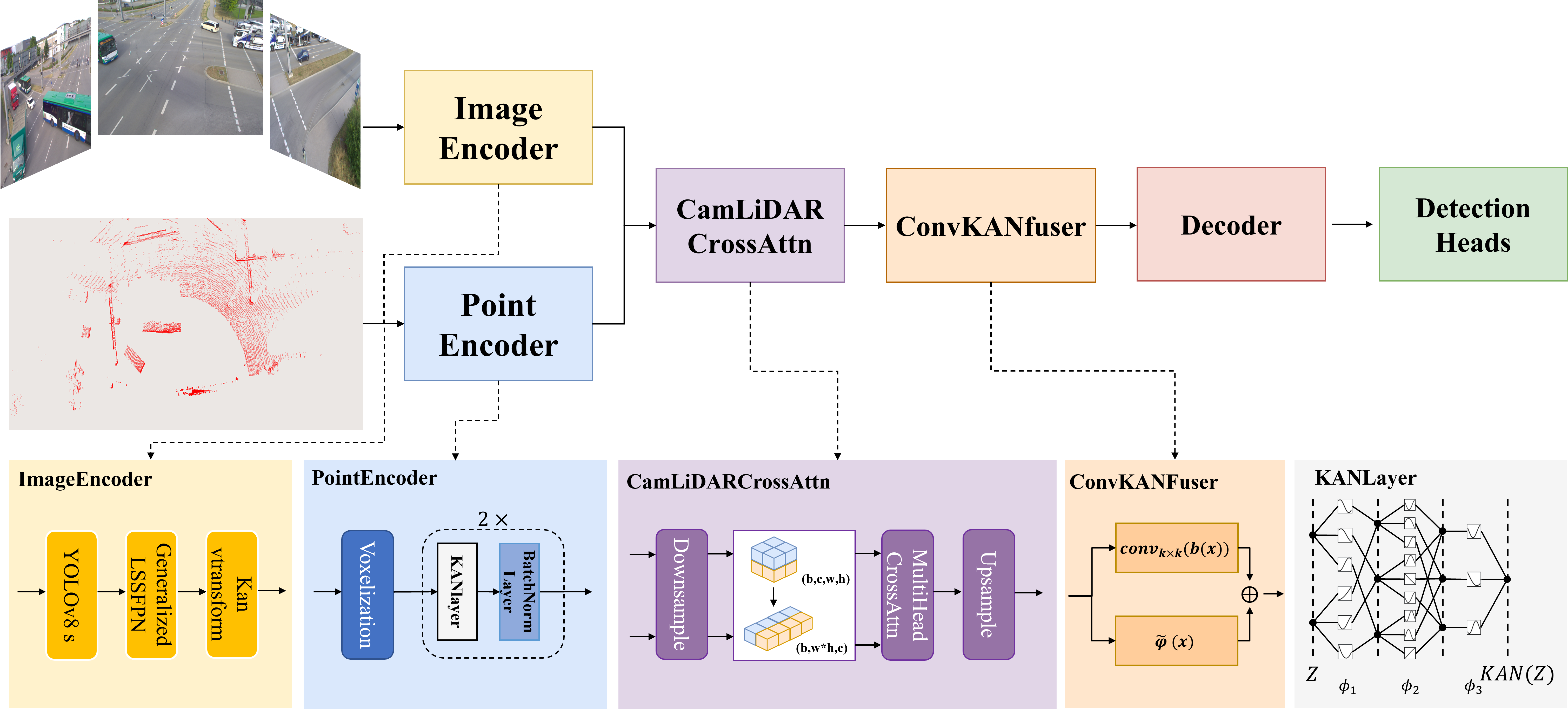}
    \caption{Model Structure of Kaninfradetr3D}
    \label{fig:Structure}
\end{figure*}

In this section, we give a complete overview of our model, Kaninfradet3D, which focuses on improving the extraction of subtle characteristics and feature fusion in roadside perception. We first describe the KAN Layers and KANConv mathematical expressions utilized later. Next, we will go into detail about the KAN-modified encoder and fuser. Finally, we present the Camera LiDAR CrossAttn module, which uses a multi-head cross-attention mechanism to compute global attention values for the mutual dependencies of the camera and LiDAR branches.

Fig. \ref{fig:Structure} illustrates the overall architecture of the model based on the roadside component of Coopdet3D. It applies an encoder to both the image branch and the LiDAR point cloud branch. The features are input into the CamLiDARCrossAttn module, where they undergo downsampling to compress the features and rearrange before being input into the MultHeadCrossAttn layer to compute cross-attention. The features are then upsampled to restore the dimensions. The ConvKANfuser is used to fuse the LiDAR and visual features, and finally, a decoder and detection head output the 3D targets.

\subsubsection{KANLayers}
This work focuses on leveraging KANs to enhance current modules. The foundational building blocks of most modern deep learning models are MLPs. MLPs essentially use multiple layers of transition matrices and activation functions to approximate complex nonlinear functions. Mathematically, this can be expressed as:

\begin{equation}
    MLP\left( \mathbf{Z} \right) = \left( W_{K - 1} \circ \sigma \circ W_{K - 2}\circ \sigma \circ \cdots \circ W_ 1 \circ \sigma \circ W_ 0~ \right)\mathbf{Z}
\end{equation}

The capacity to approximate complex functions gets better as the number of layers rises. While learning more complicated functions is theoretically possible for a large enough model, it comes at a high computational cost. This makes MLPs parameter inefficient and limits their ability to handle high-dimensional data. To address these issues, Liu et al.\cite{liu2024kan} proposed  Kolmogorov-Arnold Networks (KANs) based on the Kolmogorov-Arnold representation theorem. Similar to MLPs, KANs consist of multiple KAN layers. However, unlike MLPs, where activation functions are applied at the nodes, KANs learn activation functions on the edges. The following is a representation of KANs:

\begin{equation}
    KAN\left( \mathbf{Z} \right) = \left( {{\Phi}_{K - 1} \circ {\Phi}_{K - 2} \circ \cdots \circ {\Phi}_{1} \circ {\Phi}_{0}} \right)\mathbf{Z}
\end{equation}

where ${\Phi}_i$ represents the i-th layer of the KANs. For a KAN layer with $n_i$ dimensional input and $n_{i+1}$ dimensional output, ${\Phi}$ contains $n_i\times n_{i+1}$ learnable activation functions, similar to the edges in MLPs. The operation between the i-th and the $i+1$-th layers in a $K$-layer KANs is as follows:

\begin{equation}
    x_{i + 1,p} = {\sum\limits_{p = 1}^{n_{i}}\,}{\overset{\sim}{x}}_{l,q,p} = {\sum\limits_{p = 1}^{n_{i}}\,}\phi_{i,q,p}\left( x_{i,p} \right),q = 1,\cdots,n_{i + 1}
\end{equation}

It can also be represented as a matrix as follows:

\begin{equation}
    \mathbf{Z}_{i + 1} = \underset{{\Phi}_{i}}{\underbrace{\begin{pmatrix}
{\phi_{i,1,1}( \cdot )} & {\phi_{i,1,2}( \cdot )} & \cdots & {\phi_{i,1,n_{i}}( \cdot )} \\
{\phi_{i,2,1}( \cdot )} & {\phi_{i,2,2}( \cdot )} & \cdots & {\phi_{i,2,n_{i}}( \cdot )} \\
 \vdots & \vdots & & \vdots \\
{\phi_{i,n_{i + 1},1}( \cdot )} & {\phi_{i,n_{i + 1},2}( \cdot )} & \cdots & {\phi_{i,n_{i + 1},n_{i}}( \cdot )}
\end{pmatrix}}}\mathbf{Z}_{i}
\end{equation}

In this case $\phi_{i,q,p}$ represents the sum of basis function b(x) and the spline function.

\begin{equation}
    \phi(x) = w_{b}b(x) + {\overset{\sim}{\phi}(x)}
\end{equation}

\begin{equation}   
{\overset{\sim}{\phi}(x)} = w_{s}~spline(x)
\end{equation}

In  \cite{liu2024kan} it is suggested to set $b(x) = SiLU(x) = x/\left( {1 + e^{- x}} \right)$. KANs use parameterized activation functions as weights, discarding the linear weight matrices used in MLPs. This allows KANs to reduce model size while maintaining performance and enhancing interpretability.

\subsubsection{Kolmogorov-Arnold Convolutions(KANConv)}
\begin{equation}
\begin{split}
    x_{ij} = {\sum\limits_{l = 1}^{c}\,}{\sum\limits_{m = 0}^{k - 1}\,}{\sum\limits_{n = 0}^{k - 1}\,}\varphi_{m,n,l}\left( X_{l,i + m,j + n} \right); & i = 1,\cdots,w - k + 1,  \\
     & j = h - k + 1
\end{split}
\end{equation}

This layer will be referred to as KANConv in this paper.

\subsubsection{KANvtransform, PointEncoder and ConKANfuser}
The vtransform module is part of the BEVFusion encoder's camera branch. It consists of three modules in order: downsample, depthnet, and dtransform. The dtransform operation is a process that unfolds features in one step through three consecutive convolution blocks. Each convolution block consists of a convolutional layer, a BatchNorm layer, and a ReLU activation function. We replaced the convolutional layer with KANConv. In this study, KANvtransform is used in the ImageEncoder to convert the visual features extracted by YOLOv8s into the BEV perspective.

\begin{figure*}
    \centering
    \includegraphics[width=0.9\linewidth]{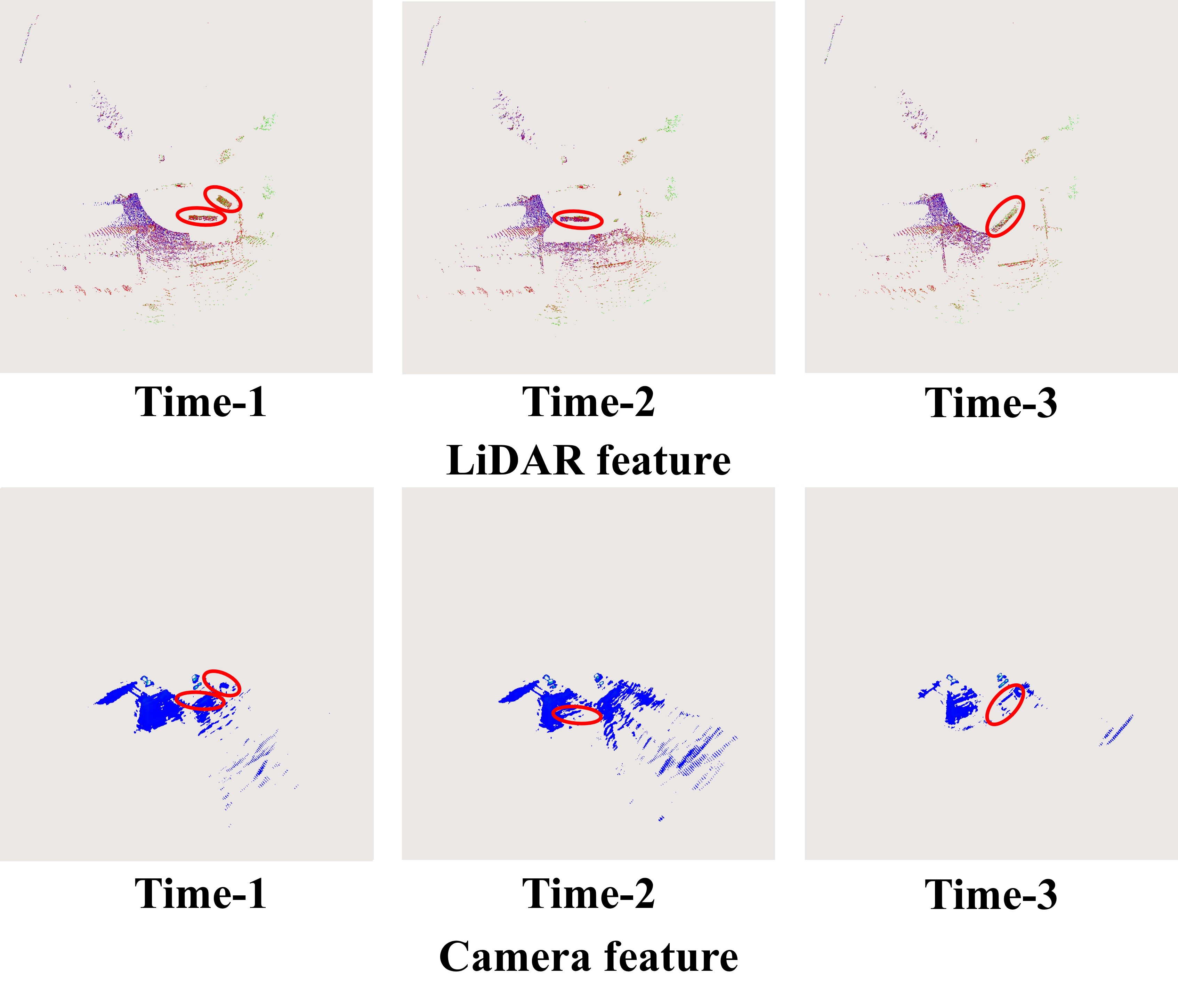}
    \caption{Visual comparison between LiDAR feature and Camera feature.The red-circled elements of the same target are easily identifiable in LiDAR, but the camera feature do not clearly separate the target from the backdrop.}
    \label{fig:feature}
\end{figure*}

The PointEncoder is based on the simplified PointNet used in PointPillarsEncoder. After modification, this module includes two consecutive Pillar Feature Nets, where the linear layers are replaced with KANLayer. The PointEncoder is used to process voxelized LiDAR point clouds. Each KANLayer is followed by a BatchNorm layer. The processed features are then scattered back to obtain the tensor $\left(b,c,w,h\right)$, where w and h indicate the feature sizes.
The ConKANfuser is located after the CrossAttn layer and includes a KANConv layer followed by BatchNorm and ReLU. It further fuses the features weighted by attention weights.

\subsection{Camera LiDAR Cross-Attention Block}

LiDAR sensors collect discrete and sparse 3D spatial features, whereas cameras capture continuous and dense 2D planar features. The fusion of these two modalities requires transforming them into a common coordinate system, thus they are both projected into the BEV space. However, directly using convolutions to fuse features does not fully utilize the information from both modalities.

Fig. \ref{fig:feature} intuitively shows the feature maps obtained after feature extraction by the ImageEncoder and PointEncoder. In the LiDAR features, the distinction between the target and the background in the center of the intersection is very clear. However, the point clouds of targets that have not yet entered the intersection are obstructed by gantries, making their features less discernible. Furthermore, with larger vehicles, a blind area exists behind them.

The features that the ImageEncoder extracted have values all around the map since images are continuous. To make this more apparent, we highlight features with values greater than $1.5\times{10}^{-3}$. The values at the targets do not exhibit any appreciable changes on the feature map. This suggests that even while there is plenty of information in the camera images, it is not being effectively extracted. Such camera feature maps struggle to assist the LiDAR information further and might even interfere with clear detection. Additionally, since the camera covers a smaller area compared to LiDAR, targets may be segmented. As a result, it is necessary to improve the extraction of picture features, and it is not recommended to fuse the extracted feature maps directly with LiDAR feature maps.

Therefore, in this study, a Camera LiDAR CrossAttn module is added after the feature extraction stage. This module is a multi-head cross-attention mechanism that calculates the dependency between LiDAR and camera features in the BEV space.
First, the feature maps of the two branches in this study are of size $(768, 768)$. Direct embedding would cause significant memory pressure. Thus, a downsampling layer compresses the features to a size of $(128, 128)$. Then, the embedded representations of the LiDAR and camera features, $E_L$ and $E_c$, are generated with dimensions $(b,128*128,c)$, where b is the batch size and c is the number of channels.

Since we want to select local regions of the image based on information in the LiDAR features, we use the LiDAR features as the query Q and the image features as the keys K and values V. Using a multi-head attention mechanism effectively stacks n different cross-attentions. The formula for splitting the multi-head attention mechanism is as follows:
\begin{equation}
    \text{MultiHead}\ (Q,K,V) = Concat\left( {\text{head}_{1},\text{head}_{2},\ldots,\text{head}_{n}} \right)W^{O}
\end{equation}
where each ${\mathrm{head}}_i$ is calculated as:
\begin{equation}
 Q_i={W_i^Q}E_{L_i},K_i={W_i^K}E_{c_i},V_i{=W_i^V}E_{c_i}, i=1,\cdots,n
\end{equation}
\begin{equation}
    {\mathrm{head}}_i=Cross\ attention(Q_i,K_i,V_i)=softmax(\frac{Q_i{K_i}^T}{\sqrt{d_k}})V_i
\end{equation}

\section{Experiments and Result Analysis}
\subsection{Experimental settings}
\textbf{Dataset}
We used the TUMTraf V2X Cooperative Perception Dataset and the TUMTraf Intersection Dataset, released by Walter Zimmer et al., for training. The TUMTraf V2X Cooperative Perception Dataset comprises 2,000 labeled point clouds and 5,000 labeled pictures, for a total of 30k 3D bounding boxes. The data was gathered utilizing multi-view high-resolution cameras and LiDARs from both vehicle and roadside perspectives. We focus exclusively on the data collected from roadside facilities. The TUMTraf Intersection Dataset includes 4.8k images and over 57.4k 3D bounding boxes. Since the test set has not been released, we use the validation set to evaluate the model results.

\textbf{Metrics}
To evaluate the accuracy of 3D bounding box detection in this study, we adopted the same evaluation metrics as those used in \cite{zimmer2023tumtraf}. The distance and number of points inside the 3D box are used to classify the targets into Hard, Moderate, and Easy classes. Less than 20 points in the box and targets farther than 50 meters are considered hard targets. Objects that are partially occluded, within a distance of 40 to 50 meters, and contain 20 to 50 points are classified as Moderate. Easy targets are those with more than 50 points, less than 40 meters distant, and no occultation. The evaluation metrics code is sourced from the TUM Traffic dataset Devkit.

\textbf{Model}
The model uses YOLOv8 s as the backbone for the vision branch and a KAN-modified PointPillar as the LiDAR backbone. Multi-head cross-attention is employed to extract relevant features between the two branches, generating feature maps of the same size as the LiDAR features. ConKANfuser then fuses the features from the multiple modalities. Subsequently, the output is processed through a decoder and detection head, with SECOND serving as the backbone.

\textbf{Implementation Details}
To facilitate easy module replacement, the model is constructed and managed using mmdet3D. We trained the model for 100 epochs on three NVIDIA GeForce RTX 4090 GPUs, taking approximately 36 hours. The model utilizes the AdamW optimizer with a learning rate of 0.0001 and a weight decay of 0.01 for regularization during the training process. The memory demand for KANLayer operations and cross-attention computations is significantly high, necessitating the use of a batch size of 1 to maintain robust training. The initial learning rate is set to $1\times{10}^{-4}$. To further stabilize the training process, we employed a Cosine Annealing strategy, starting with a warmup ratio of 0.3 to gradually increase the learning rate at the beginning of the training; as the training progresses, the learning rate is systematically decreased to ensure gradual convergence to the optimal solution.

Our training is based on the pre-trained weights released by \cite{zimmer2024tumtraf}, and YOLOv8 s also uses the same pre-trained weights as used. The modules replaced with KAN layers are trimmed from the pre-trained weights. Due to memory limitations, the training process is divided into three stages: First, we train the modules that replaced the linear layers, specifically the PointEncoder. The linear layer weights are trimmed, and the model is trained for 20 epochs. Next, we train the modified KANConv layers separately, freezing the weights of the KanLinear layers for another 20 epochs. Finally, we conduct a full model training for 60 epochs.

\subsection{Results}

In this section, we compare our proposed method with the benchmarks Coopdet3D from the TUMTraf V2X Cooperative Perception Dataset and  InfraDet3D from the TUMTraf Intersection Dataset . Both Coopdet3D and InfraDet3D have numerous variations, including interchangeable backbone plugins such as YOLO s and SwinT, PointPillars and VoxelNet. They also offer options for applying different data modalities and fusion stages. Due to the unavailability of the test set, our model is evaluated on the validation set. As this study does not involve vehicle-side data, the Coopdet3D models mentioned below refer to those conducted on the roadside end, without integrating vehicle-side data.

Table \ref{tab:table1} displays the results of all current benchmarks and Kaninfradet3D on the TUMTraf Intersection Dataset. Notably, in both the south 1 and south 2 fields of view, the radar modality of Coopdet3D outperforms the fused camera and radar results. This observation highlights the motivation behind our study. In Kaninfradet3D, by effectively integrating valuable information, the camera-LiDAR fusion surpasses the performance of the LiDAR-only branch in Kaninfradet3D. Furthermore, Kaninfradet3D achieves significant improvements of +9.87 mAP and +10.64 mAP over Coopdet3D in the two different camera views. These improvements are mainly attributed to the pre-fusion of features using cross-attention and the enhanced feature extraction capabilities of the Encoder module provided by the KAN network.

\begin{table}[]
    \centering
    \caption{Evaluation results of Kaninfradet3D and benchmark models on TUMTraf Intersection Dataset}
    \begin{tabular}{l|c|c|c|c}
    \hline
      \hline
        \textbf{Method} & \textbf{FOV} & \textbf{Modality} & \textbf{Fusion Level} & \bm{$mAP_{\bm{3D}}$}\ensuremath{\uparrow} \\
        \hline
        PointPillars & {--} & Lidar &  Early Fusion & 62.11 \\
    Coopdet3D & South1 & Lidar  & Deep Fusion & 69.47 \\
    Infradet3D & South1 & Camera, Lidar  & Late Fusion & 44.55 \\
    Coopdet3D & South1 & Camera, Lidar  & Deep Fusion & 66.75
    \\
    Coopdet3D & South2 & Lidar  & Deep Fusion & 69.94 \\
    Infradet3D & South2 & Camera, Lidar  & Late Fusion & 37.06 \\
    Coopdet3D & South2 & Camera, Lidar  & Deep Fusion & 66.89 \\
    Kaninfradet3D & South1 & Camera, Lidar  & Deep Fusion & \textbf{76.62} \\
    Kaninfradet3D & South2 & Camera, Lidar  & Deep Fusion & \textbf{77.53} \\
    \hline
      \hline
    \end{tabular}
    \label{tab:table1}
\end{table}

Additionally, we evaluated the model's performance on the TUMTraf V2X Cooperative Perception Dataset under different difficulty levels, which is depicted in Table \ref{tab:table2}. On this dataset, KanInfraDet3D showed less pronounced improvements compared to the previous dataset, achieving +1.96 mAP and +3.05 mAP in the Easy and Moderate categories, respectively. However, it did not outperform Coopdet3D in the Hard category. This discrepancy might be attributed to the varying quality of datasets, particularly the specific scene occlusions. Overall, these results demonstrate that our model provides accurate perception and exhibits stable performance. They also validate the value and potential of KAN Layers in 3D perception tasks.

\begin{table}
    \centering
    \caption{Evaluation results of Kaninfradet3D and benchmark models on the infrastructure part of TUMTraf V2X Cooperative Perception Dataset}
    \begin{tabular}{l|c|c|c|c|c}
    \hline
    \hline
    \multirow{2}*{\textbf{Method}} & \multirow{2}*{\textbf{Modality}} & \multicolumn{4}{|c}{\bm{$mAP_{\bm{3D}}$}\ensuremath{\uparrow}} \\
    \cline{3-6}
         ~ & ~ &  \textbf{Easy} & \textbf{Mid}& \textbf{Hard} & \textbf{Avg} \\
         \hline
        Coopdet3D(infra) & Camera & 31.19&	46.73	&40.42	&35.04 \\
        Coopdet3D(infra)	&Lidar& 	86.17&	88.07&	75.73	&84.88 \\
        Coopdet3D(infra)	&Camera, Lidar &	87.99	&89.09	&\textbf{81.69}	&87.01 \\
        Kaninfradet3D	&Camera, Lidar &	\textbf{89.95} &	\textbf{92.14}	&77.04	&\textbf{88.41} \\
        \hline
        \hline
    \end{tabular}
    \label{tab:table2}
\end{table}

\subsection{Ablation Study on Camera-LiDAR Attention}

\begin{figure*}
    \centering
    \includegraphics[width=0.9\linewidth]{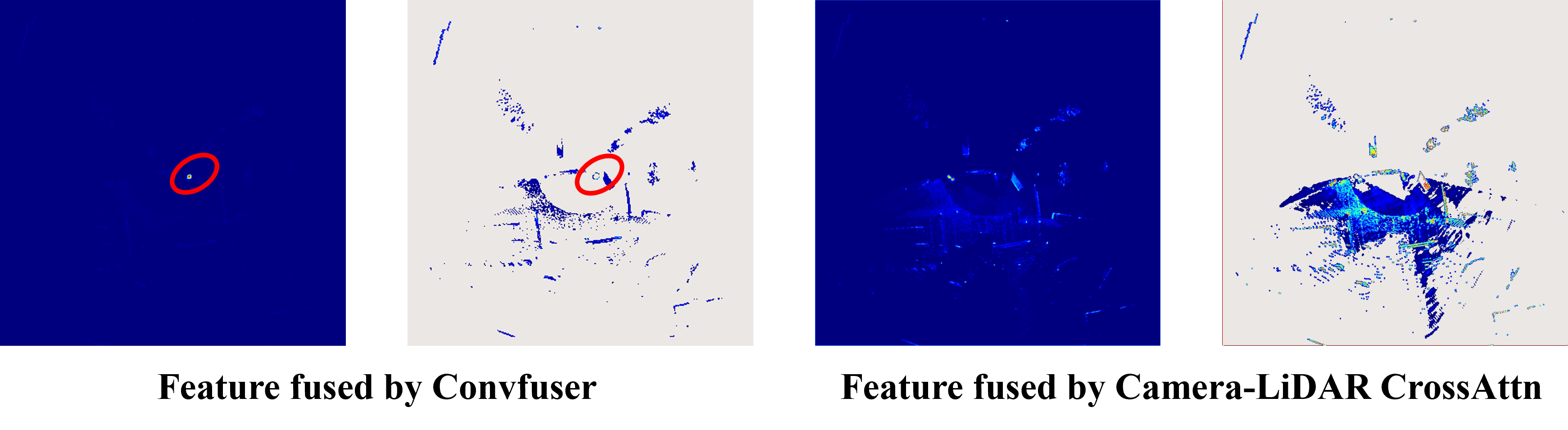}
    \caption{Comparison of features after fusion of Convfuser and Camera-LiDAR CrossAttn. The part circled in red is the abnormally fused camera feature. The gray background image is a version with the background removed for clearer display.}
    \label{fig:compare}
\end{figure*}

We compared the features before and after applying Camera-LiDAR CrossAttn with those that did not apply Camera-LiDAR CrossAttn in order to visually validate the effectiveness of the Camera-LiDAR CrossAttn module in feature fusion. The comparison is quite clear. The visual characteristics mainly affected a tiny area in front of the camera, and the fused features were primarily reliant on LiDAR data prior to the addition of Camera-LiDAR CrossAttn. In some cases, the fused features showed highly uneven distribution, with extremely high values in regions marked by red circles in the Fig. \ref{fig:compare}, indicating an abnormal concentration of camera features. This concentration hindered the effective learning of other features in the global context.

After integrating the Camera-LiDAR CrossAttn module, the camera features were more evenly distributed across the LiDAR features rather than being abnormally concentrated in one area. Such abnormal fusion of camera features was common in the original module, but the Camera-LiDAR CrossAttn effectively resolved this issue. After successfully integrating the camera features, not only were the targets with strong LiDAR reflection signals highlighted, but the features of smaller targets in the background, such as pedestrians and motorcycles, also became more distinguishable in the feature map.

In Table \ref{tab:my_label}, we evaluated the impact of improvements made to each module, with the enhancement from the Camera-LiDAR CrossAttn being the most significant. This demonstrates the critical role of the Camera-LiDAR CrossAttn in camera-LiDAR fusion. The model achieved mAP scores of 75.44 and 75.54 in the two views of the TUMTraf Intersection Dataset, which are the closest to the performance of the complete Kaninfradet3D model.

\subsection{Ablation Study on KANvtransform, PointEncoder and ConKANfuser}

To evaluate the effectiveness of KANs in the feature extraction and fuser modules, we separately assessed the modules after incorporating KAN modifications. In the PointEncoder, KAN Layers replace the linear layers, while in the Kanvtransform and ConvKANfuser, Conv2D layers are replaced with KANConv, so these two were grouped. The results in \ref{tab:my_label} show that the improvement in the PointEncoder was the least significant, while the enhancements from Kanvtransform and ConvKANfuser were more substantial, with increases of +4.48 mAP and +4.66 mAP, respectively. Overall, the incorporation of KANs produces favorable results, whether in the form of linear layers or convolutional layers.

The most significant improvement came from the inclusion of cross-attention, indicating that at this stage, refining fusion methods yields greater benefits than improving feature extraction. Our model achieves optimal performance by combining these three modules.

\begin{table}
\setlength\tabcolsep{4pt}
    \centering
    \begin{tabular}{cccc|cc}
    \hline
    \hline
     
\multirow{2}*{\textbf{\makecell[c]{Point\\Encoder}}} & \multirow{2}*{\textbf{\makecell[c]{Kanv\\transform}}}&\multirow{2}*{\textbf{\makecell[c]{Conv\\KANfuser}}} &\multirow{2}*{\textbf{\makecell[c]{Camera\\LiDARAttn}}} & \multicolumn{2}{c}{\bm{$mAP_{\bm{3D}}$}\ensuremath{\uparrow}} \\
\cline{5-6}
      &  &  &  & \textbf{South1}  & \textbf{South2}  \\
    \hline 
        &          &            &         & 66.75 & 66.89\\
        \checkmark &            &         &        &70.05 &70.37\\
         & \checkmark & \checkmark &  & 71.23 &  71.55\\
         &      &          &\checkmark & 75.44&  75.54\\
      \checkmark & \checkmark & \checkmark &\checkmark & \textbf{76.62}&  \textbf{77.53}\\
        \hline
        \hline
    \end{tabular}
    \caption{Ablation Study on KANs modified module and cross-attention module}
    \label{tab:my_label}
\end{table}

\section{Conclusions}

This paper investigates the current challenges in roadside perception, such as the lack of effective methods and models, inadequate fine-grained feature extraction in existing fusion frameworks, and suboptimal fusion performance among different branches. To tackle these issues, we introduce Kolmogorov-Arnold Networks and three modified KAN modules derived from the KAN Layer. Two feature encoders, a multi-head cross-attention module, a KAN-enhanced convolutional fuser, a decoder, and a detection head are all part of the suggested network.

KAN Layers, which take the place of conventional linear layers in the PointEncoder, improve the encoder's capacity to recognize and comprehend high-dimensional characteristics. KANConv enhances the learning of complicated features by substituting the original Conv2D layers in the ImageEncoder and ConvKANfuser. Prior to the fuser, the Camera-LiDARAttn module is added. Its purpose is to pre-fuse features from different branches by calculating multi-head cross-attention on them based on their dependencies. This successfully lessens the detrimental effects on the fusion process of regionally high values in camera characteristics. By utilizing these innovative designs, Kaninfradet3D exhibits exceptional accuracy in roadside 3D perception challenges, offering fresh perspectives on roadside 3D object detection.

Even though our model works exceptionally well on the training dataset, additional validation on a broader range of datasets is required due to the recent rapid development of roadside 3D perception datasets. Additionally, there are many details regarding the integration of KANs that warrant further discussion, including the number of added layers, analysis of model parameters, and a deeper exploration of their interpretability in perception tasks. We will subsequently focus on these aspects, continuing to research and optimize the modules to update the model. 

\addtolength{\textheight}{-12cm}   





\section*{ACKNOWLEDGMENT}
The research is supported by Southeast University Interdisciplinary Research Program for Young Scholars (Grant No. 2024FGC1006), and Shenzhen Technology Program Project (Grant No. SGDX20230821095159012).



\end{document}